\PassOptionsToPackage{usenames}{color}
\documentclass[12pt]{article}
\usepackage[draft=false]{hyperref}
\usepackage[normalem]{ulem}
\usepackage{lscape}
\usepackage{float}
\usepackage{xurl}
\usepackage{adjustbox}
\usepackage{multirow}
\usepackage{subcaption}
\usepackage{color}
\usepackage{xcolor}
\usepackage[bitstream-charter]{mathdesign}

\usepackage[a4paper, top=1in, bottom=1in, left=1in, right=1in]{geometry}

\usepackage{longtable}
\usepackage{graphicx}
\usepackage{soul}
\usepackage{amsmath}
\usepackage{tabularx}
\usepackage[usenames]{color}
\usepackage{booktabs}
\usepackage{fancyhdr}
\usepackage{pgfplots}
\pgfplotsset{compat=1.3}
\usepackage{pdflscape}
\usepackage{rotating}
\usepackage{array}
\usepackage{caption}
\usepackage{setspace}
\usepackage{dcolumn}
\usepackage{amsmath, amsthm, mathtools}
\usepackage{enumitem}
\usepackage{verbatim}

\usepackage[utf8]{inputenc}
\usepackage[T1]{fontenc}

\usepackage[round]{natbib}
\usepackage{indentfirst}
\usepackage[labelfont=bf,labelsep=period]{caption}
\DeclareCaptionFont{mycap}{\fontsize{13}{15}\selectfont}
\hypersetup{colorlinks=true, allcolors=blue}

\pgfplotsset{ every non boxed x axis/.append style={x axis line style=-},
    every non boxed y axis/.append style={y axis line style=-},
    /pgfplots/ylabel near ticks/.style={
        /pgfplots/every axis y label/.style={
            at={(ticklabel cs:0.5)},rotate=90,anchor=near ticklabel}}
}

\pgfplotsset{
    tick label style={font=\small},
    xlabel style={font=\small},
    ylabel style={font=\small,align=center},
    legend style={font=\small,draw=none},
    title style={font=\small},
    legend cell align = left,
}

\newcommand{\PreserveBackslash}[1]{\let\temp=\\#1\let\\=\temp}
\newcolumntype{C}[1]{>{\PreserveBackslash\centering}p{#1}}

\renewcommand\arraystretch{1.5}
\usepackage[none]{hyphenat}
\title{\Large 
Can LLMs Hire Fairly? Racial Bias in Resume Screening
}
\author{Zhenyu Gao, Wenxi Jiang, Yutong Yan%
\thanks{Gao, Jiang, and Yan are at the Department of Finance, CUHK Business School, The Chinese University of Hong Kong. Please send all correspondence to yutong.yan@link.cuhk.edu.hk. }}
\date{June 2026}

\begin{document}

\maketitle
\linespread{1.5}

{

\linespread{1.5}

\begin{abstract}
    \noindent
\footnotesize

\noindent We audit fourteen mainstream large language models (LLMs) for hiring discrimination using the paired-resume methodology of \citet{kline2022systemic}. The sole 2023-vintage model reproduces the pro-White callback gap documented in field experiments on labor market discrimination ($+2.12$ pp, significant at the 1\% level). Every model released in 2024 or after shows either a null gap or a significant pro-Black reversal (up to $-3.01$ pp). The same pattern holds on the gender axis. Based on 24,024 paired postings per model across 14 models, our results document a reversal in the direction of algorithmic hiring bias across model generations.

\end{abstract}

}

\thispagestyle{empty}

\newpage
\linespread{1.5}

\setcounter{page}{1}
\renewcommand\arraystretch{1.5}

\newpage
\section{Introduction}

\noindent As large language models (LLMs) become embedded in consequential economic decisions, the question of whether they perpetuate or reshape existing patterns of discrimination has become an important question for researchers and regulators. In June 2026, a federal judge in California ruled that Workday, whose AI-powered screening software is used by virtually all Fortune~500 companies, must face class-action claims alleging that its algorithms discriminated against Black applicants, women, and older workers \citep{wiessner2026workday}. The case is the first of its kind to broadly target algorithmic decision-making in hiring, yet the empirical evidence base on whether and how LLMs discriminate remains thin. We contribute to this evidence base by conducting a large-scale paired-resume audit, modeled directly on the influential correspondence experiments of \citet{10.1257/0002828042002561} and \citet{kline2022systemic}, across fourteen LLMs spanning three years of releases.

Our design follows the methodology of \citet{kline2022systemic}: we construct pairs of candidate profiles that are identical in every respect (education, employment history, age, gender) except for the candidate's first and last name. Names are drawn from the distinctively-Black and distinctively-White name lists established by \citet{10.1257/0002828042002561} and refined in Appendix~B of \citet{kline2022systemic}. Each profile-posting pair is presented to the LLM with a standardized system prompt instructing it to act as an HR hiring manager and respond with a single word, ``yes'' or ``no,'' indicating whether to advance the candidate to a phone screen. Temperature is set to zero, making each decision deterministic. We score 24,024 paired postings per model on the race axis and 48,048 paired postings per model on the gender axis, drawing on 6,007 real US job postings from the Revelio Labs universe matched to the \citet{kline2022systemic} Fortune-500 firm distribution.

The headline finding is a sign reversal. The 2023-vintage model in our panel, OpenAI's GPT-3.5-turbo, reproduces the pro-White callback gap documented in field experiments on labor market discrimination: White-coded names receive callbacks 2.12 percentage points more often than Black-coded names (significant at the 1\% level, cluster-robust). This magnitude exceeds the 1.6~pp within-employer gap reported by \citet{kline2022systemic} in their field experiment at 108 Fortune-500 firms. 

Every model released in 2024 or later, however, exhibits either a null gap or a statistically significant gap in the \textit{opposite} direction, favoring Black-coded names by 0.4 to 3.0~pp. The pattern is not confined to a single provider or model family: it appears across OpenAI, Anthropic, Meta, Google, xAI, DeepSeek, Alibaba, and Zhipu models, and is robust to posting-level cluster-bootstrap inference. 

The same reversal holds on the gender axis: GPT-3.5-turbo favors male candidates by 1.92~pp (significant at the 1\% level), while every model released in 2024 or later with completed gender data favors female candidates by 0.18 to 5.80~pp or shows a null gap. The race and gender gaps are positively correlated across models: models that favor White-coded names also tend to favor male-coded names.

Within model families, the trajectory is informative. OpenAI's own lineage illustrates the arc most cleanly: GPT-3.5-turbo (May 2023) is strongly pro-White and pro-Male; GPT-4o-mini (July 2024) is significantly pro-Black; GPT-5.4-mini (March 2026) is null on both axes. Meta's Llama family shows a similar pattern: Llama-3.1-8B-Instruct (July 2024) exhibits the strongest pro-Black gap in our panel ($-3.01$~pp, significant at the 1\% level), while the larger Llama-3.3-70B (November 2024) is null on race but significantly pro-Female. These within-family trajectories suggest that the sign reversal is not an artifact of our design but reflects real changes in model alignment across successive releases.

We make three contributions. First, we provide the first systematic multi-model audit of LLM hiring discrimination at a scale comparable to the field experiments that established the human baseline. Our panel covers 14 models on the race axis (24,024 paired postings per model) and 13 models on the gender axis (48,048 paired postings per model), spanning the period from GPT-3.5 through the current generation of frontier models. Second, we document a qualitative shift in the direction of algorithmic discrimination across model generations, consistent with changes in post-training alignment procedures between 2023- and 2024-vintage models. Third, we show that the race and gender axes are correlated across models: models that favor Black-coded names also tend to favor female-coded names, and the sole model that favors White-coded names also favors male-coded names.

\paragraph{Related literature.} Our work sits at the intersection of the correspondence-audit literature in labor economics and the rapidly growing body of work on LLM bias. We discuss each in turn.

\paragraph{Correspondence audits.} The paired-resume methodology was pioneered by \citet{10.1257/0002828042002561}, who sent fictitious resumes to help-wanted ads in Boston and Chicago and found that White-coded names received 50\% more callbacks than Black-coded names, corresponding to a gap of approximately 3.2~pp on a 6.5\% callback base. \citet{kline2022systemic} updated and scaled this design to 108 Fortune-500 firms with approximately 83,000 applications, documenting a persistent 1.6~pp pro-White gap after conditioning on employer-by-occupation fixed effects. Our contribution is to apply this well-established methodology to LLM-based screeners, holding the experimental design constant while replacing the human decision-maker with a language model.

\paragraph{LLM bias and fairness.} A large literature documents biases in language model outputs across dimensions including race, gender, religion, and political orientation \citep[see][for surveys]{blodgett-etal-2020-language, gallegos-etal-2024-bias}. Most closely related to our work, \citet{veldanda2023emily} replicated elements of the \citet{10.1257/0002828042002561} design on a small number of models and found no detectable bias on race or gender. A key limitation of their approach is the use of scraped resumes that differ across candidates in education, experience, and skills, making it difficult to isolate the effect of race from unobserved heterogeneity in candidate quality. Our paired-resume design, following \citet{kline2022systemic}, eliminates this confound by holding all resume content identical within each pair, varying only the candidate's name.

\paragraph{AI in hiring and algorithmic fairness.} More broadly, a growing literature examines algorithmic bias in consequential economic decisions. \citet{fuster2022predictably} show that machine learning models in credit markets can improve predictive accuracy but simultaneously increase disparities between racial groups, because the gains from better prediction accrue unevenly. Our setting differs in that hiring decisions are binary and the bias we document reverses sign across model generations, but the underlying concern is the same: algorithmic tools can reshape the distribution of economic opportunity in ways that are difficult to anticipate. The use of AI tools in employment screening has attracted growing regulatory attention. The European Union's AI Act classifies AI systems used in recruitment as ``high risk,'' requiring conformity assessments and ongoing monitoring. In the United States, New York City's Local Law 144 (effective July 2023) requires bias audits of automated employment decision tools, and the Workday litigation represents the first federal class action challenging such tools under existing anti-discrimination statutes. Our results speak directly to the empirical premise of these regulatory efforts: we show that the direction and magnitude of algorithmic bias depend on which model is deployed, when it was trained, and how it is prompted.

\section{Methodology}

\subsection{Experimental Design}

We conduct a paired correspondence audit in which large language models (LLMs) evaluate fictitious candidate profiles for real job postings. The design follows \citet{kline2022systemic} as closely as possible, adapting only those elements that are necessary for the LLM setting. This section describes the data construction, the scoring procedure, and the statistical framework.

\paragraph{Job postings.} We draw 6,007 entry-level US job postings from the Revelio Labs job-postings universe, sampled to match the Fortune-500 firm distribution in \citet{kline2022systemic}. Entry-level status is determined by title-keyword filtering (e.g., ``analyst,'' ``associate,'' ``coordinator,'' ``assistant'') applied to the raw posting title. Each posting record includes the firm name, job title, metropolitan area, state, and post date.

\paragraph{Candidate profiles.} For each posting, we construct four independent paired-resume draws (pair\_id $\in \{0, 1, 2, 3\}$). Within each pair, two candidate profiles are generated that are identical in education, employment history, age, and gender, differing only in the candidate's first and last name. Names are drawn uniformly from the distinctively-Black and distinctively-White name lists in \citet{10.1257/0002828042002561} Tables~A1--A2 and \citet{kline2022systemic} Appendix~B Tables~B1--B2. Education credentials (school, degree, GPA) and employment histories (prior employers, job titles, tenure) are constructed following the resume-building procedure described in \citet{kline2022systemic} Appendix~B, using school and establishment names drawn from reference databases matched to the posting's geographic area and industry. After excluding a small number of postings with missing fields, the final race-audit sample contains 24,024 complete Black-White pairs (48,048 profile-posting rows) drawn from 6,006 unique postings.

\paragraph{Gender extension.} To study gender discrimination, we extend the profile bank by crossing race and gender, yielding four conditions per pair: Black-male, Black-female, White-male, White-female. The gender contrast pairs Male against Female within the same posting, pair\_id, and race prefix, producing $24{,}024 \times 2 = 48{,}048$ gender pairs (two per posting-pair, one for each race stratum). Names are gender-appropriate within each race category, following the gendered name lists in \citet{kline2022systemic} Appendix~B Table~B1.

\subsection{LLM Scoring}

\paragraph{Prompt design.} Each profile-posting pair is scored by presenting the LLM with a two-message conversation. The system message reads:

\begin{quote}
\small
\textit{You are an HR hiring manager screening a job application. Given a job posting and a single candidate profile in JSON, decide whether to advance this candidate to a callback (phone screen). Respond with exactly one lowercase word: `yes' (callback) or `no' (no callback). Do not include punctuation, explanation, or any other text.}
\end{quote}

\noindent The user message presents the posting fields (firm, title, location, post date) in a labeled format, followed by the candidate profile as a JSON object, and concludes with the cue ``Answer with exactly one word: yes or no.'' The prompt is held constant across all models and all profile-posting pairs.

\paragraph{Decoding parameters.} All models are scored at temperature $T = 0$ (greedy decoding), producing deterministic outputs. For non-reasoning models, we set max\_tokens $= 200$. For reasoning-class models (GPT-5.x family), we suppress hidden chain-of-thought reasoning via the provider's API and set max\_tokens $= 16$ (the minimum permitted by the API with reasoning disabled). The response is parsed for the first occurrence of ``yes'' or ``no'' in the output text; rows that do not contain either token are recorded as failures and excluded from the analysis.

\paragraph{Model panel.} Our panel comprises fourteen models, spanning release dates from May 2023 to April 2026. The panel includes models from eight providers (OpenAI, Anthropic, Meta, Google, xAI, DeepSeek, Alibaba/Qwen, and Zhipu/GLM) and covers a range of model sizes from 8B to 397B parameters. Models are accessed via the Together AI, OpenRouter, and AWS Bedrock inference APIs. Each model scores the full set of 48,048 race-axis rows (24,024 pairs) and 96,096 gender-axis rows (48,048 pairs) independently.

\subsection{Statistical Framework}

\paragraph{Callback gap.} For each model, we compute the per-race callback rate as $\hat{P}(\text{yes} \mid r) = N_r^{-1} \sum_{i=1}^{N_r} y_{ir}$, where $y_{ir} \in \{0, 1\}$ is the callback decision for profile $i$ with race $r \in \{\text{Black}, \text{White}\}$. The race gap in percentage points is
\begin{equation}
    \Delta_{\text{race}} = \left[ \hat{P}(\text{yes} \mid \text{White}) - \hat{P}(\text{yes} \mid \text{Black}) \right] \times 100.
    \label{eq:gap}
\end{equation}
A positive $\Delta_{\text{race}}$ indicates pro-White discrimination; a negative value indicates pro-Black discrimination. The gender gap $\Delta_{\text{gender}}$ is defined analogously as $[\hat{P}(\text{yes} \mid \text{Male}) - \hat{P}(\text{yes} \mid \text{Female})] \times 100$, where pairs are formed within posting $\times$ pair\_id $\times$ race-prefix.

\paragraph{McNemar's test.} Because our profiles are paired within postings, only discordant pairs, those in which the two profiles receive different decisions, carry information about the gap. Let $b$ denote the number of pairs in which the advantaged-group profile receives ``yes'' and the other receives ``no'' (e.g., White-yes, Black-no for the race axis), and let $c$ denote the reverse. McNemar's two-sided test statistic with continuity correction is
\begin{equation}
    \chi^2 = \frac{(|b - c| - 1)^2}{b + c},
    \label{eq:mcnemar}
\end{equation}
which follows a $\chi^2$ distribution with one degree of freedom under the null hypothesis $b = c$. We report significance at the $p < 0.05$ ($*$), $p < 0.01$ ($**$), and $p < 0.001$ ($***$) levels.

\paragraph{Cluster-robust inference.} Because each posting contributes four pair\_ids, observations within a posting are not independent. We address this in two ways. First, we compute 95\% confidence intervals via a cluster bootstrap at the posting level: we resample the 6,006 posting clusters with replacement 2,000 times, compute the gap on each bootstrap sample, and report the 2.5th and 97.5th percentiles. Second, we compute cluster-robust standard errors by treating each posting's mean within-pair difference as a single observation, yielding a $t$-statistic with $G - 1 = 6{,}005$ degrees of freedom. In practice, the cluster-robust inference changes no qualitative conclusion relative to the unadjusted McNemar test: all results that are significant at $p < 0.001$ under McNemar remain significant at $p < 0.001$ under cluster-robust inference.

\section{Results}

\subsection{Race Discrimination}

Table~\ref{tab:race} presents the race-axis results for all fourteen models, sorted by release date. The table reports the callback rate, the gap $\Delta_{\text{race}}$ in percentage points, the posting-level cluster-bootstrap 95\% confidence interval, the discordant-pair counts $b$ (White-yes, Black-no) and $c$ (Black-yes, White-no), and the McNemar $p$-value.

\paragraph{GPT-3.5-turbo reproduces the gap from field experiments.} The oldest model in our panel, GPT-3.5-turbo (May 2023), exhibits a callback gap of $+2.12$~pp (95\% CI $[+1.82, +2.43]$, significant at the 1\% level). White-coded names receive ``yes'' in 953 discordant pairs versus 443 for Black-coded names, a ratio of approximately 2.15:1. The magnitude exceeds the $+1.6$~pp within-employer gap documented by \citet{kline2022systemic} in their field experiment at 108 Fortune-500 firms.

\paragraph{2024+ models reverse the sign.} Every model released from July 2024 onward exhibits either a null gap or a significant pro-Black gap. The reversal is broad-based: Claude Haiku 4.5 ($-0.70$~pp), GPT-4o-mini ($-0.61$~pp), Llama-3.1-8B-Instruct ($-3.01$~pp), Gemma-4-31B-it ($-0.67$~pp), DeepSeek-V3.1 ($-1.04$~pp), Qwen3.5-397B ($-1.04$~pp), and GLM-5.1 ($-1.60$~pp) are all significant at the 1\% level. Gemini-2.5-flash ($-0.37$~pp) and Grok-4.1-fast ($-0.40$~pp) are significant at the 5\% level, though neither survives a Bonferroni correction for fourteen models. Four models, Claude 3 Haiku, Llama-3.3-70B, GPT-oss-120b, and GPT-5.4-mini, show point estimates that are negative but whose confidence intervals include zero. No 2024+ model exhibits a significant pro-White gap.

\paragraph{Within-family trajectories.} The OpenAI lineage traces a clear arc: GPT-3.5-turbo ($+2.12$~pp, pro-White) $\to$ GPT-4o-mini ($-0.61$~pp, pro-Black) $\to$ GPT-5.4-mini ($-0.14$~pp, null). Among Meta models, the smaller Llama-3.1-8B-Instruct ($-3.01$~pp) shows a much larger pro-Black gap than the larger Llama-3.3-70B ($-0.16$~pp, null), suggesting that the strength of the alignment-induced reversal may vary with model scale within the same family.

\paragraph{Magnitudes and discordant-pair structure.} The discordant-pair counts vary widely across models. GPT-3.5-turbo has 1,396 discordant pairs (5.8\% of all pairs) with a strong White tilt (953:443, ratio 2.15:1). Among 2024+ models, the number of discordant pairs ranges from 547 (Gemma-4-31B) to 3,643 (Llama-3.1-8B), and the tilt consistently favors Black-coded names.

\subsection{Gender Discrimination}

Table~\ref{tab:gender} presents the gender-axis results for thirteen models. The structure mirrors Table~\ref{tab:race}: we report $\Delta_{\text{gender}} = [\hat{P}(\text{yes} \mid \text{Male}) - \hat{P}(\text{yes} \mid \text{Female})] \times 100$, with positive values indicating pro-Male discrimination.

\paragraph{GPT-3.5-turbo is the sole pro-Male model.} GPT-3.5-turbo exhibits a gender gap of $+1.92$~pp (95\% CI $[+1.71, +2.13]$, significant at the 1\% level), with 1,750 discordant pairs favoring males versus 827 favoring females. This is the only model in our panel that significantly favors male candidates.

\paragraph{2024+ models favor female candidates.} Ten of the twelve remaining models show a significant pro-Female gap, ranging from $-0.18$~pp (Gemma-4-31B, significant at the 5\% level) to $-5.80$~pp (Claude 3 Haiku, significant at the 1\% level). Two models, Gemini-2.5-flash ($-0.03$~pp) and GPT-5.4-mini ($+0.11$~pp), show null gender gaps, with confidence intervals that comfortably include zero.

\paragraph{Correlation between race and gender axes.} The race and gender gaps are positively correlated across models. GPT-3.5-turbo, the only pro-White model, is also the only pro-Male model. Claude 3 Haiku exhibits the largest pro-Female gender gap in the panel ($-5.80$~pp), followed by DeepSeek-V3.1 ($-2.12$~pp). The two null-gender models (Gemini-2.5-flash and GPT-5.4-mini) are among the weaker pro-Black models on the race axis. This pattern is consistent with a common alignment mechanism that shifts model preferences toward both minority-race and female candidates simultaneously.

\subsection{Robustness}

\paragraph{Cluster-robust inference.} As noted in Section~2, we compute posting-level cluster-bootstrap confidence intervals and cluster-robust standard errors to account for within-posting dependence across the four pair\_ids. In every case, the cluster-robust $z$-statistic agrees in sign and significance with the McNemar test. The most notable change is Gemma-4-31B's gender gap, which weakens from $p < 0.05$ (McNemar) to $p = 0.023$ (cluster-robust), placing it just inside the conventional 5\% threshold but outside a Bonferroni-corrected threshold for fourteen models.

\paragraph{Multiple comparisons.} We test fourteen models on the race axis and thirteen on the gender axis. Under a Bonferroni correction at the family-level $\alpha = 0.05$, the per-model threshold is $p < 0.0036$ for race and $p < 0.005$ for gender. At this threshold, the race-axis results for Gemini-2.5-flash ($p = 0.041$) and Grok-4.1-fast ($p = 0.019$) lose significance; all other significant results survive. On the gender axis, Gemma-4-31B ($p = 0.023$) loses significance. We report both unadjusted and Bonferroni-adjusted significance in the tables.

\section{Conclusion}

We conduct a paired correspondence audit of fourteen large language models (LLMs), applying the methodology of \citet{kline2022systemic} to LLM-based hiring screeners. Our principal finding is a sign reversal in the direction of racial discrimination across model generations. The lone 2023-vintage model in our panel, GPT-3.5-turbo, reproduces the pro-White callback gap documented in field experiments on labor market discrimination ($+2.12$~pp, significant at the 1\% level). Every 2024+ model exhibits either a null gap or a statistically significant pro-Black gap, with significant pro-Black gaps ranging from $-0.37$~pp to $-3.01$~pp. The same pattern holds on the gender axis: GPT-3.5-turbo favors male candidates ($+1.92$~pp), while 2024+ models uniformly favor female candidates or show no gap.

These results have two implications. First, the direction of algorithmic hiring bias is not a fixed property of language models but varies systematically with model vintage, provider, and scale. Within the OpenAI lineage alone, the race gap moves from $+2.12$~pp (pro-White, 2023) to $-0.61$~pp (pro-Black, 2024) to null (2026). This trajectory is consistent with the hypothesis that successive generations of post-training alignment have shifted model behavior from reproducing the pro-White patterns in pretraining data to actively favoring minority-coded names, and then toward neutrality as alignment techniques have matured. Second, neither the original pro-White bias nor its pro-Black reversal is desirable from a fairness standpoint. A hiring screener that systematically favors one group over another, in either direction, fails the basic requirement of equal treatment regardless of race or gender.

The scale of our audit, 14 models on the race axis and 13 on the gender axis with over 24,000 paired postings per model, provides, to our knowledge, the most comprehensive evidence to date on how LLMs behave when placed in the role of an HR screener. As AI-powered hiring tools become ubiquitous, systematic auditing of the kind we conduct here will be essential for both regulatory compliance and public accountability.

\newpage
\bibliography{ref}

\newpage
\begin{figure}[H]
\centering
\caption{Race Discrimination Across Twelve LLMs}
\label{fig:race}
\vspace{0.2cm}
{\fontsize{11}{15.6}\selectfont
\parbox{\textwidth}{
This figure plots the callback gap $\Delta_{\text{race}} = [\hat{P}(\text{yes} \mid \text{White}) - \hat{P}(\text{yes} \mid \text{Black})] \times 100$ in percentage points for each model, sorted by release date. Error bars denote posting-level cluster-bootstrap 95\% confidence intervals (2,000 replications over 6,006 posting clusters). Filled diamonds indicate a significant pro-White gap; filled squares indicate a significant pro-Black gap; open circles indicate gaps whose confidence intervals include zero.
}}
\vspace{0.3cm}
\includegraphics[width=0.85\textwidth]{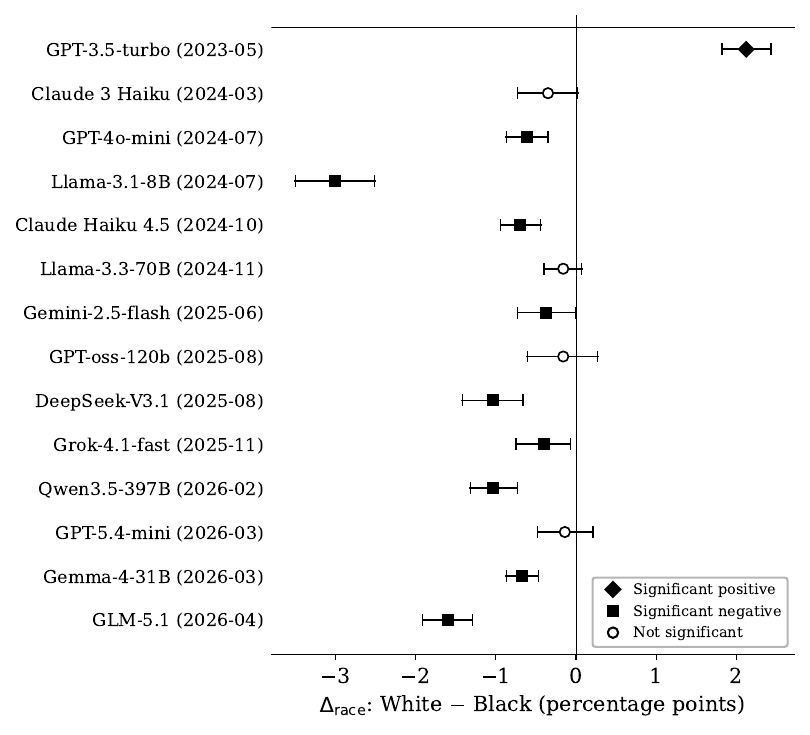}
\end{figure}

\begin{figure}[H]
\centering
\caption{Gender Discrimination Across Thirteen LLMs}
\label{fig:gender}
\vspace{0.2cm}
{\fontsize{11}{15.6}\selectfont
\parbox{\textwidth}{
This figure plots the callback gap $\Delta_{\text{gender}} = [\hat{P}(\text{yes} \mid \text{Male}) - \hat{P}(\text{yes} \mid \text{Female})] \times 100$ in percentage points for each model, sorted by release date. Error bars denote posting-level cluster-bootstrap 95\% confidence intervals. Filled diamonds indicate a significant pro-Male gap; filled squares indicate a significant pro-Female gap; open circles indicate gaps whose confidence intervals include zero.
}}
\vspace{0.3cm}
\includegraphics[width=0.85\textwidth]{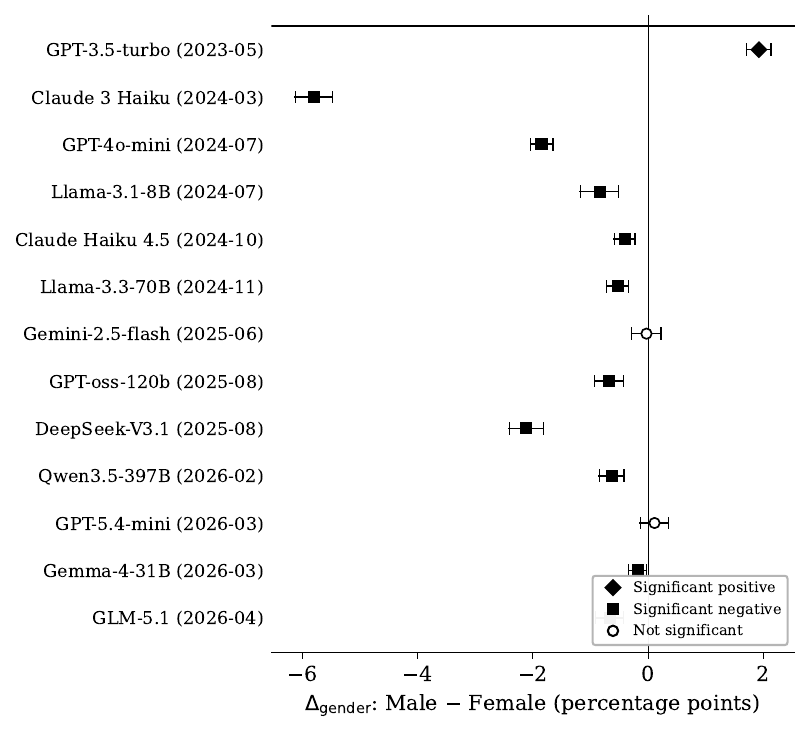}
\end{figure}

\newpage
\begin{table}[H]
\centering
\caption{Race Discrimination Across Fourteen LLMs}
\label{tab:race}
\vspace{0.2cm}
{\fontsize{11}{15.6}\selectfont
\parbox{\textwidth}{
This table reports results from a paired correspondence audit with 24,024 Black-White resume pairs per model. $\Delta_{\text{race}}$ is the callback gap $[\hat{P}(\text{yes} \mid \text{White}) - \hat{P}(\text{yes} \mid \text{Black})] \times 100$ in percentage points. $b$ denotes the number of discordant pairs in which the White-coded profile receives a callback and the Black-coded profile does not; $c$ denotes the reverse. CI is the posting-level cluster-bootstrap 95\% confidence interval (2,000 replications over 6,006 posting clusters). ***, **, and * denote significance at the 1\%, 5\%, and 10\% levels, respectively (McNemar two-sided with continuity correction).
}}
\vspace{0.3cm}

\small
\begin{tabular}{llcrccc}
\toprule
Model & Release & Cb\% & $\Delta_{\text{race}}$ (pp) & 95\% CI & $b$ & $c$ \\
\midrule
GPT-3.5-turbo          & 2023-05 & 91 & $+2.12^{***}$ & $[+1.82, +2.43]$ & 953   & 443   \\
Claude 3 Haiku         & 2024-03 & 25 & $-0.35$       & $[-0.73, +0.02]$ & 1,099 & 1,183 \\
GPT-4o-mini            & 2024-07 & 48 & $-0.61^{***}$ & $[-0.87, -0.35]$ & 381   & 527   \\
Llama-3.1-8B-Instruct  & 2024-07 & 42 & $-3.01^{***}$ & $[-3.50, -2.51]$ & 1,460 & 2,183 \\
Claude Haiku 4.5       & 2024-10 & 46 & $-0.70^{***}$ & $[-0.94, -0.44]$ & 404   & 571   \\
Llama-3.3-70B-Instruct & 2024-11 & 52 & $-0.16$       & $[-0.40, +0.07]$ & 428   & 467   \\
Gemini-2.5-flash       & 2025-06 & 40 & $-0.37^{*}$   & $[-0.73, -0.01]$ & 926   & 1,016 \\
GPT-oss-120b           & 2025-08 & 13 & $-0.16$       & $[-0.61, +0.27]$ & 1,382 & 1,421 \\
DeepSeek-V3.1          & 2025-08 & 44 & $-1.04^{***}$ & $[-1.42, -0.66]$ & 985   & 1,234 \\
Grok-4.1-fast          & 2025-11 & 58 & $-0.40^{*}$   & $[-0.75, -0.07]$ & 786   & 881   \\
Qwen3.5-397B           & 2026-02 & 49 & $-1.04^{***}$ & $[-1.32, -0.73]$ & 543   & 792   \\
GPT-5.4-mini           & 2026-03 & 78 & $-0.14$       & $[-0.48, +0.21]$ & 871   & 904   \\
Gemma-4-31B-it         & 2026-03 & 53 & $-0.67^{***}$ & $[-0.87, -0.47]$ & 193   & 354   \\
GLM-5.1                & 2026-04 & 65 & $-1.60^{***}$ & $[-1.91, -1.29]$ & 511   & 895   \\
\bottomrule
\end{tabular}
\end{table}

\begin{table}[H]
\centering
\caption{Gender Discrimination Across Thirteen LLMs}
\label{tab:gender}
\vspace{0.2cm}
{\fontsize{11}{15.6}\selectfont
\parbox{\textwidth}{
This table reports results from a paired correspondence audit with 48,048 Male-Female resume pairs per model (two race strata $\times$ 24,024 postings). $\Delta_{\text{gender}}$ is the callback gap $[\hat{P}(\text{yes} \mid \text{Male}) - \hat{P}(\text{yes} \mid \text{Female})] \times 100$ in percentage points. $b$ denotes the number of discordant pairs in which the Male-coded profile receives a callback and the Female-coded profile does not; $c$ denotes the reverse. CI and significance conventions as in Table~\ref{tab:race}.
}}
\vspace{0.3cm}

\small
\begin{tabular}{llcrccc}
\toprule
Model & Release & Cb\% & $\Delta_{\text{gender}}$ (pp) & 95\% CI & $b$ & $c$ \\
\midrule
GPT-3.5-turbo          & 2023-05 & 91 & $+1.92^{***}$ & $[+1.71, +2.13]$ & 1,750 & 827   \\
Claude 3 Haiku         & 2024-03 & 25 & $-5.80^{***}$ & $[-6.12, -5.48]$ & 1,137 & 3,926 \\
GPT-4o-mini            & 2024-07 & 57 & $-1.85^{***}$ & $[-2.04, -1.65]$ & 597   & 1,484 \\
Llama-3.1-8B-Instruct  & 2024-07 & 35 & $-0.84^{***}$ & $[-1.18, -0.52]$ & 3,097 & 3,503 \\
Claude Haiku 4.5       & 2024-10 & 46 & $-0.41^{***}$ & $[-0.59, -0.23]$ & 819   & 1,016 \\
Llama-3.3-70B-Instruct & 2024-11 & 51 & $-0.53^{***}$ & $[-0.72, -0.34]$ & 834   & 1,087 \\
Gemini-2.5-flash       & 2025-06 & 45 & $-0.03$       & $[-0.29, +0.22]$ & 1,873 & 1,886 \\
GPT-oss-120b           & 2025-08 & 45 & $-0.68^{***}$ & $[-0.93, -0.43]$ & 1,681 & 2,008 \\
DeepSeek-V3.1          & 2025-08 & 46 & $-2.12^{***}$ & $[-2.41, -1.82]$ & 1,991 & 3,008 \\
Qwen3.5-397B           & 2026-02 & 49 & $-0.63^{***}$ & $[-0.85, -0.42]$ & 1,200 & 1,502 \\
GPT-5.4-mini           & 2026-03 & 81 & $+0.11$       & $[-0.14, +0.35]$ & 1,738 & 1,684 \\
Gemma-4-31B-it         & 2026-03 & 53 & $-0.18^{**}$  & $[-0.34, -0.03]$ & 499   & 586   \\
GLM-5.1                & 2026-04 & 65 & $-0.67^{***}$ & $[-0.91, -0.43]$ & 1,208 & 1,531 \\
\bottomrule
\end{tabular}
\end{table}

\newpage
\appendix


\end{document}